%% file: main.tex
\documentclass[10pt,twocolumn,letterpaper]{article}

\usepackage[pagenumbers]{cvpr} % To force page numbers, e.g. for an arXiv version
\usepackage{hyperref}
\input{preamble}

\definecolor{cvprblue}{rgb}{0.21,0.49,0.74}
\usepackage{amsmath}
\usepackage{amssymb}
\usepackage{mathtools}

\newcommand{\bx}{\mathbf{x}}

\newcommand{\bz}{\mathbf{z}}

\newcommand{\bI}{\mathbf{I}}

\def\paperID{*****} % *** Enter the Paper ID here
\def\confName{CVPR}
\def\confYear{2024}

\title{VideoCrafter1: Open Diffusion Models for High-Quality Video Generation}

\author{
Haoxin Chen$^{1}$\thanks{Equal contribution.} \quad
Menghan Xia$^{1}$\footnotemark[1] \quad
Yingqing He$^{1,2}$\footnotemark[1] \quad
Yong Zhang$^{1}$\footnotemark[1]\quad
Xiaodong Cun$^1$\footnotemark[1] \quad   \\
Shaoshu Yang$^{1,3,4}$  \quad
Jinbo Xing$^{1,5}$\quad
Yaofang Liu$^{1,6}$\quad
Qifeng Chen$^{2}$\quad \\
Xintao Wang$^{1}$\thanks{Corresponding author.}\quad
Chao Weng$^{1}$\quad
Ying Shan$^{1}$\quad
\\\\
\textsuperscript{\rm 1}~Tencent AI Lab \quad
\textsuperscript{\rm 2}~Hong Kong University of Science and Technology \\
\textsuperscript{\rm 3}~Center for Research on Intelligent Perception and Computing, CASIA \quad \\
\textsuperscript{\rm 4}~School of Artificial Intelligence, UCAS\\ 
\textsuperscript{\rm 5}~The Chinese University of Hong Kong
\quad
\textsuperscript{\rm 6}~City University of Hong Kong
\\\\
{{Project page: {\url{https://ailab-cvc.github.io/videocrafter}}}} \\
{{GitHub: {\url{https://github.com/AILab-CVC/VideoCrafter}}}}
}

\begin{document}

\twocolumn[{
\maketitle
\begin{center}
    \captionsetup{type=figure}
    \includegraphics[width=0.95\textwidth]{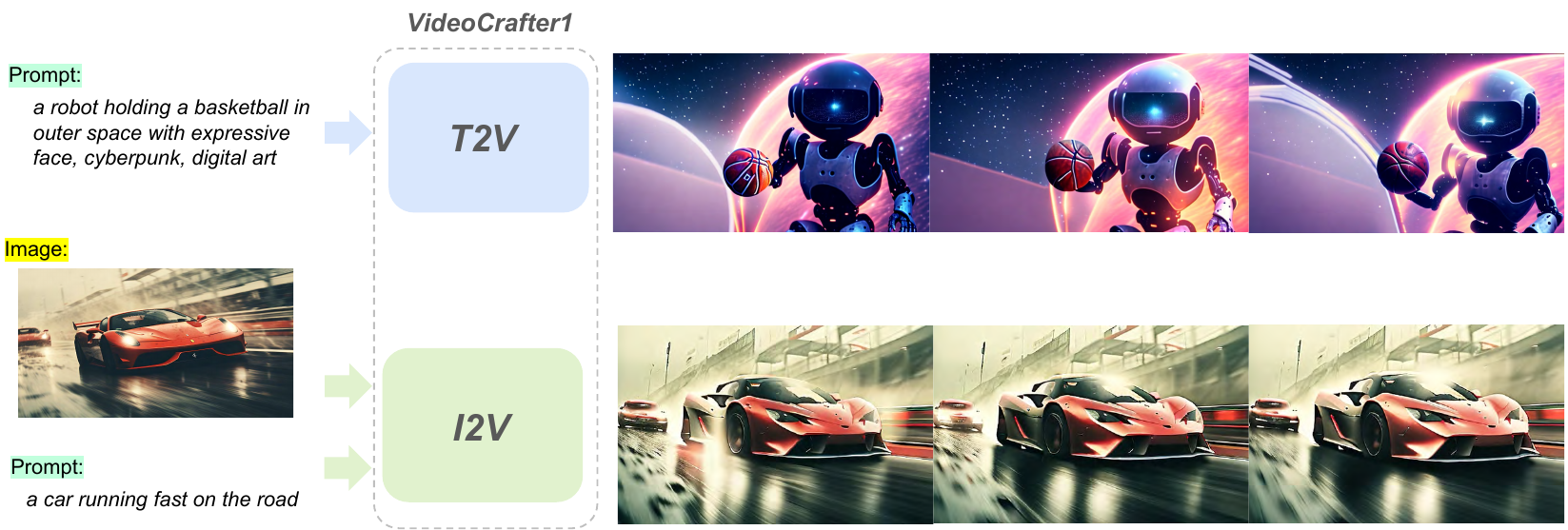}
    \captionof{figure}{ 
   We have open-sourced two diffusion models for video generation in VideoCrafter1. The Text-to-Video (T2V) model takes a text prompt as input and generates a video accordingly. On the other hand, the Image-to-Video (I2V) model accepts either an image, a text prompt, or both as input for video generation.
    }
\end{center}
}]

{
  \renewcommand{\thefootnote}%
    {\fnsymbol{footnote}}
  \footnotetext[1]{~Equal contribution.} \footnotetext[2]{~Corresponding author.}
}

\input{tex/0-abstract}

\input{tex/1-introduction}

\input{tex/2-related-works}

\input{tex/3-method}

\input{tex/4-experiment}

\input{tex/5-conclusion}

{
    \small
    \bibliographystyle{ieeenat_fullname}
    \bibliography{main}
}

\end{document}

%% file: preamble.tex
\usepackage[dvipsnames]{xcolor}

%% file: tex/0-abstract.tex
\begin{abstract}
Video generation has increasingly gained interest in both academia and industry. Although commercial tools can generate plausible videos, there is a limited number of open-source models available for researchers and engineers. In this work, we introduce two diffusion models for high-quality video generation, namely text-to-video (T2V) and image-to-video (I2V) models. T2V models synthesize a video based on a given text input, while I2V models incorporate an additional image input. Our proposed T2V model can generate realistic and cinematic-quality videos with a resolution of $1024 \times 576$, outperforming other open-source T2V models in terms of quality. The I2V model is designed to produce videos that strictly adhere to the content of the provided reference image, preserving its content, structure, and style. This model is the first open-source I2V foundation model capable of transforming a given image into a video clip while maintaining content preservation constraints. We believe that these open-source video generation models will contribute significantly to the technological advancements within the community.

\end{abstract}

%% file: tex/1-introduction.tex
\section{Introduction}
\label{sec:intro}

With the rapid development of generative models, particularly diffusion models~\cite{ho2020denoising,song2020denoising}, numerous breakthroughs have been achieved in fields such as image generation~\cite{saharia2022photorealistic,ramesh2022hierarchical,podell2023sdxl,dai2023emu,gafni2022make,yu2022scaling,chang2023muse,chen2023pixart} and video generation~\cite{singer2022make,ho2022imagenvideo,he2022latent,blattmann2023align,wang2023modelscope}, as well as in recognition and detection tasks~\cite{chen2023diffusiondet,nguyen2023diffusion}.
The most well-known open-source text-to-image (T2I) generative model is Stable Diffusion (SD)~\cite{rombach2022high}, which produces plausible results.
Subsequently, its enhanced version, SDXL~\cite{podell2023sdxl}, was released, offering improved concept composition and image quality.
Another notable T2I open-source model is IF~\cite{IF2023}, a cascaded model that operates on pixels rather than latent features.
Regarding text-to-video (T2V) models, Make-A-Video~\cite{singer2022make} and Imagen Video~\cite{ho2022imagenvideo} are cascaded models, while most other works, such as LVDM~\cite{he2022latent}, Magic Video~\cite{zhou2022magicvideo}, ModelScope~\cite{wang2023modelscope}, and Align your Latents~\cite{blattmann2023align}, are SD-based models.
These models extend the SD framework to videos by incorporating temporal layers to ensure temporal consistency among frames.
The spatial parameters are inherited from the pretrained SD UNet.

The flourishing of successful T2I models and advancements in downstream tasks can be largely attributed to the open-source environment within the community.
SD serves as a critical foundation, as it is trained on a vast collection of text-image pairs using immense computing power.
The cost associated with this is often prohibitive for most academic research groups.
In contrast, in the field of T2V, Make-A-Video~\cite{singer2022make} and Imagen Video~\cite{ho2022imagenvideo} demonstrate promising video results, but neither of them are open-sourced.
Several startups, such as Gen-2~\cite{Gen2}, Pika Labs~\cite{pikalab}, and Moonvalley~\cite{moonvalley}, can generate high-quality videos, but their models remain inaccessible to researchers for further exploration.

Currently, several open-source T2V models exist, \textit{i.e.,} ModelScope~\cite{wang2023modelscope}, Hotshot-XL~\cite{hotshotxl}, AnimateDiff~\cite{guo2023animatediff}, and Zeroscope V2 XL~\cite{zeroscope}.
The released ModelScope model can only generate videos with a resolution of $256 \times 256$, and the image quality is unsatisfactory.
Zeroscope V2 XL improves its visual quality by tuning it on a small set of videos, but flickers and visible noise persist in its generated videos.
Hotshot-XL aims to extend SDXL into a video model and produce a gif with 8 frames and a resolution of $512\times 512$.
AnimateDiff proposes to combine the temporal module with the spatial module of a LORA SD model. 
Since the temporal module is trained on Webvid-10M, the results of the original T2V model of AnimateDiff are poor. 
The combination with a high-quality LORA model can generate high-quality videos. 
% Although the results of the original T2V model are poor, the combination can yield high-quality videos. 
However, the scope is restricted by the LORA model in terms of style and concept composition ability.
There is still a lack of an open-source generic T2V foundation model capable of generating high-resolution and high-quality videos.

Recently, Pika Labs and Gen-2 released their image-to-video (I2V) models, aiming to animate a given image with a prompt while preserving its content and structure.
Such a technique is still in its early stages, as the generated motion is limited, and there are usually visible artifacts.
The only open-source generic I2V foundation model, I2VGen-XL~\cite{I2VGen-XL}, is released in ModelScope.
This model uses image embedding to replace text embedding for tuning a pretrained T2V model.
However, it does not satisfy the content-preserving constraints. The generated videos match the semantic meaning in the given image but do not strictly follow the reference content and structure.
Hence, there is an urgent need for a good I2V model in the open-source community.

In this work, we introduce two diffusion models for high-quality video generation: one for text-to-video (T2V) generation and the other for image-to-video (I2V) generation.
The T2V model builds upon SD 2.1 by incorporating temporal attention layers into the SD UNet to capture temporal consistency.
We employ a joint image and video training strategy to prevent concept forgetting.
The training dataset comprises LAION COCO 600M~\cite{LAION-COCO}, Webvid10M~\cite{Bain21}, and a 10M high-resolution collected video dataset.
The T2V model can generate videos with a resolution of $1024 \times 576$ and a duration of $2$ seconds.
The I2V model, on the other hand, is based on a T2V model and accepts both text and image inputs.
The image embedding is extracted using CLIP~\cite{cherti2023reproducible} and injected into the SD UNet through cross attention~\cite{ye2023ip-adapter}, similar to the injection of text embeddings.
The I2V model is trained on LAION COCO 600M and Webvid10M.
By releasing these models, we aim to make a significant contribution to the open-source community, enabling researchers and practitioners to build upon our work and further advance the field of video generation.

Our contributions can be summarized as follows:
\begin{itemize}
    \item We introduce a text-to-video model capable of generating high-quality videos with a resolution of $1024 \times 576$ and cinematic quality. The model is trained on 20 million videos and 600 million images.
    \item We present an image-to-video model, the first open-source generic I2V model that can strictly preserve the content and structure of the input reference image while animating it into a video. This model allows for both image and text inputs.
\end{itemize}

%% file: tex/2-related-works.tex
\section{Related Works}
\label{sec:related-works}
%--------------------------------------

\input{fig_tex/fig_demo1}

Diffusion models (DMs)~\cite{sohl2015deep,ho2020denoising,song2021score} have recently shown unprecedented capability in content generation field, especially in text-to-image (T2I) generation~\cite{ramesh2022hierarchical,nichol2022glide,saharia2022photorealistic,gu2022vector,rombach2022high,balaji2022ediffi,he2023scalecrafter,zhang2023real,podell2023sdxl}. Following the success of T2I DMs, Video Diffusion Models (VDMs) are proposed to model the spatial and temporal distribution of videos under the condition of text prompts (T2V). The first VDM~\cite{ho2022video} utilizes a space-time factorized U-Net to model low-resolution videos in pixel space, which is trained jointly on image and video data. To generate high-definition videos, Imagen-Video~\cite{ho2022imagenvideo} introduces an effective cascaded paradigm of DMs with $\mathbf{v}$-prediction parameterization method. To promote the computational efficiency, subsequent studies~\cite{wang2023modelscope,he2022latent,blattmann2023align,zhou2022magicvideo,wang2023lavie} mainly take the way of transferring T2I knowledge to T2V generation~\cite{singer2022make,luo2023videofusion} and learning DMs in latent space. Most recently, Zhang \etal~\cite{zhang2023show1} highlight the issue of heavy computational cost in pixel-based VDM~\cite{ge2023preserve} and poor text-video alignment in latent-based VDM, proposing a hybrid-pixel-latent VDM framework to address these issues.

Although T2V models can generate high-quality videos, they only accept text prompts as semantic guidance, which can be verbose and may not accurately reflect users' intentions. Similar to adding controls in T2I models~\cite{li2023gligen,zhang2023adding,mou2023t2i,ye2023ipadapter,shi2023instancebooth}, introducing conditional controls in T2V DMs has increasingly attracted researchers' attention. Gen-1~\cite{esser2023structure} and Make-Your-Video~\cite{xing2023make} integrate structure control into VDMs by concatenating frame-wise depth map with input noise sequences for video editing, while other control conditions, including pose~\cite{ma2023follow,zhang2023controlvideo} and canny edge~\cite{zhang2023controlvideo,khachatryan2023text2video} are also investigated.
However, visual conditions in VDMs, such as RGB images, remain under-explored. Most recently, image condition is examined in Seer~\cite{gu2023seer} and VideoComposer~\cite{wang2023videocomposer} for text-image-to-video synthesis, by keeping first-frame latent clean and concatenating image embedding with noise in channel dimension, respectively. Nevertheless, they either focus on the curated domain, \ie, indoor objects~\cite{gu2023seer}, or fail to generate temporally coherent frames and realistic motions~\cite{wang2023videocomposer} due to insufficient semantic understanding of the input image. Although DragNUWA~\cite{yin2023dragnuwa} further introduce trajectory control into image-to-video generation, which can only mitigate the unrealistic-motion issue to some extent.
Moreover, some recent close-sourced text-to-video diffusion models~\cite{singer2022make,molad2023dreamix,li2023videogen} or auto-regressive models~\cite{villegas2023phenaki,yu2023magvit} successfully demonstrate their extension applicability to image-to-video synthesis. However, their results rarely adhere to the input image condition and suffers from the unrealistic temporal variation issue. While we build our model upon text-conditioned VDMs to leverage their rich dynamic prior for animating open-domain images, incorporating tailored designs for better conformity and semantic understanding of the input image.

%% file: fig_tex/fig_demo1.tex
\begin{figure*}[t]
  \centering
   \includegraphics[width=0.86\linewidth]{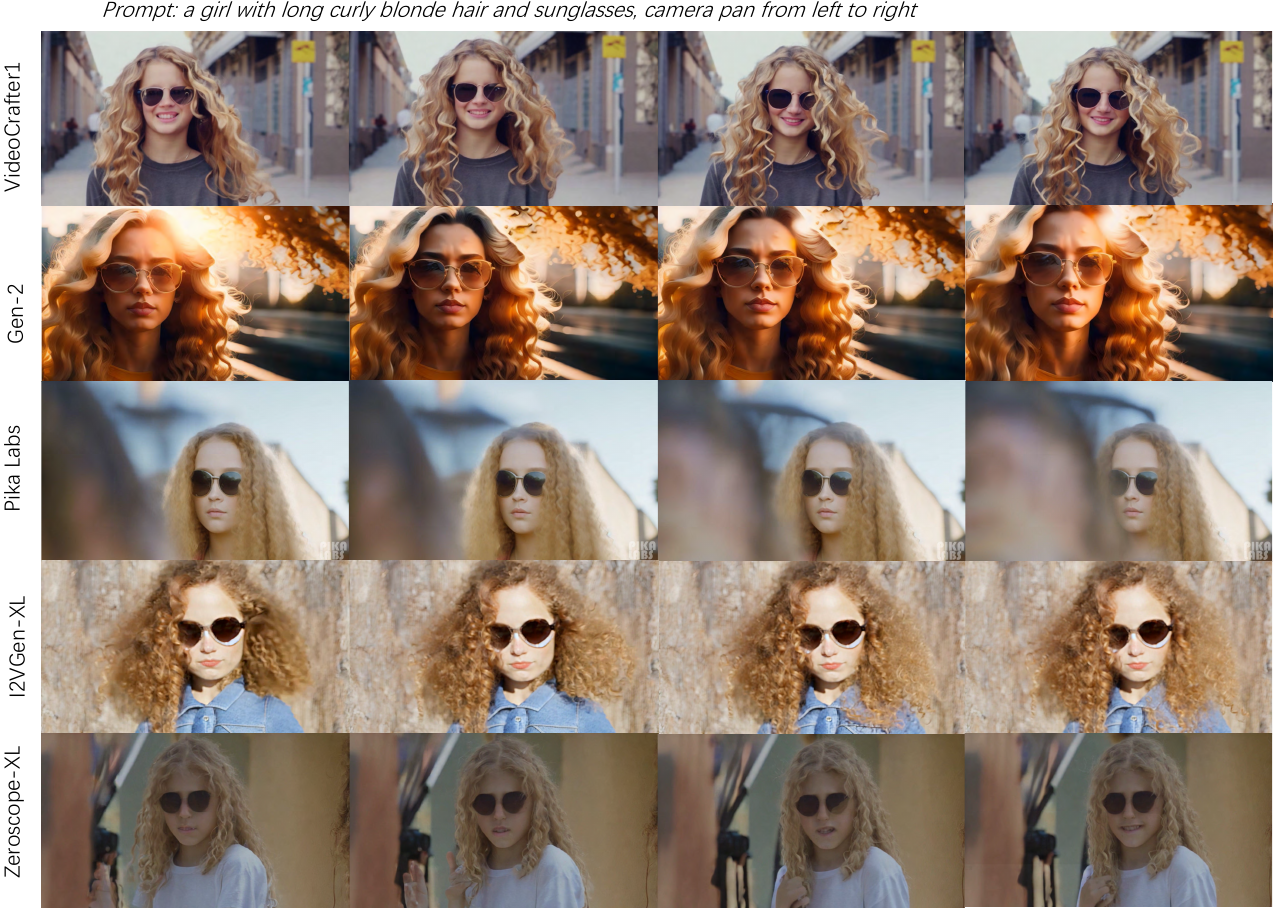}
   \includegraphics[width=0.86\linewidth]{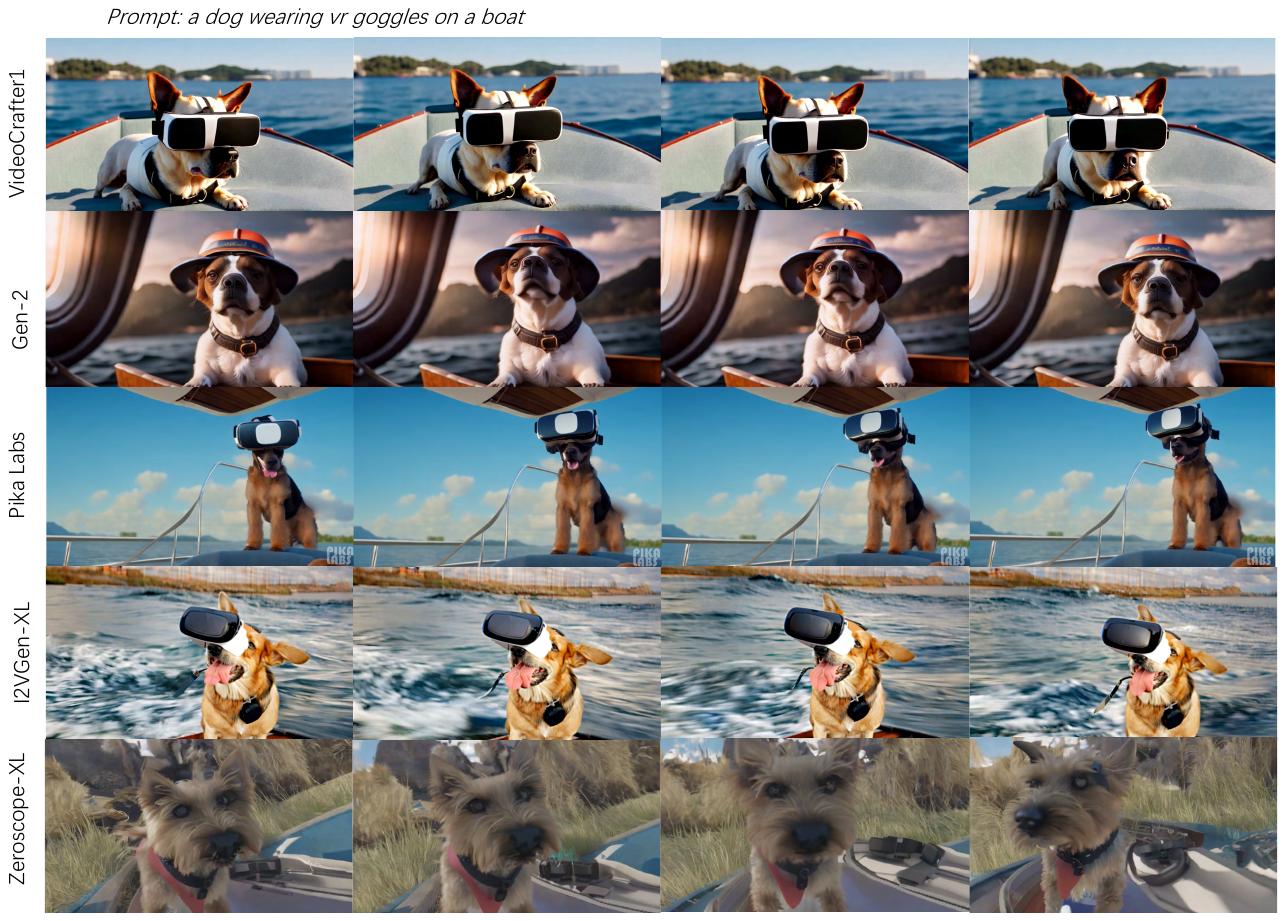}
   \caption{Visual comparison with Gen-2, Pika Labs, I2VGen-XL, and Zeroscope-XL.}
   \label{fig:demo1}
\end{figure*}

%% file: tex/3-method.tex
\section{Methodology}
\label{sec:method}
%--------------------------------------

\subsection{VideoCrafter1: Text-to-Video Model}
\label{subsec:t2v}
%--------------------------------------

\input{fig_tex/fig_frame_t2v}

\paragraph{Structure Overview.}
The VideoCrafter T2V model is a Latent Video Diffusion Model (LVDM)~\cite{he2022latent} consisting of two key components: a video VAE and a video latent diffusion model, as illustrated in Fig.~\ref{fig:framework-t2v}.
The Video VAE is responsible for reducing the sample dimension, allowing the subsequent diffusion model to be more compact and efficient.
First, the video data $\bx_0$ is fed into the VAE encoder $\mathcal{E}$ to project it into the video latent $\bz_0$, which exhibits a lower data dimension with a compressed video representation.
Then, the video latent can be projected back into the reconstructed video $\bx_0^\prime$ via the VAE decoder $\mathcal{D}$.
We adopt the pretrained VAE from the Stable Diffusion model to serve as the video VAE and project each frame individually without extracting temporal information.
After obtaining the video latent $\bz_0$, the diffusion process is performed on $\bz_0$ via:

\begin{align}
q(\bz_{1:T} | \bz_0) &\coloneqq \prod_{t=1}^T q(\bz_t | \bz_{t-1} ),\\
q(\bz_t|\bz_{t-1}) & \coloneqq \mathcal{N}(\bz_t;\sqrt{1-\beta_t}\bz_{t-1},\beta_t \bI),
\label{eq:forwardprocess}
\end{align}
where $T$ is the number of diffusion timesteps, and $\beta_t$ is the noise level at timestep $t$.
Thus, we can obtain a series of noisy video latents $\bz_t$ at arbitrary timesteps $t$.

To perform the denoising process, a denoiser U-Net is learned to estimate the noise in the input noisy latent, which will be discussed in the next section.
After the progressive denoising process, the latent sample transitions from noisy to clean, and it can finally be decoded by the VAE decoder into a generated video in the pixel space.

\paragraph{Denoising 3D U-Net.}
As illustrated in Fig.\ref{fig:framework-t2v}, the denoising U-Net is a 3D U-Net architecture consisting of a stack of basic spatial-temporal blocks with skip connections.
Each block comprises convolutional layers, spatial transformers (ST), and temporal transformers (TT), where
\begin{align}
\mathrm{ST} &= \mathrm{Proj}_{\mathrm{in}} \circ (\mathrm{Attn}_{\text{self}} \circ \mathrm{Attn}_{\text{cross}} \circ \mathrm{MLP}) \circ \mathrm{Proj}_{\text{out}}, \\
\mathrm{TT} &= \mathrm{Proj}_{\mathrm{in}} \circ (\mathrm{Attn}_{\text{temp}} \circ \mathrm{Attn}_{\text{temp}} \circ \mathrm{MLP}) \circ \mathrm{Proj}_{\text{out}}.
\end{align}
The controlling signals of the denoiser include semantic control, such as the text prompt, and motion speed control, such as the video fps.
We inject the semantic control via the cross-attention:
% $$, with

\begin{align}
\text{Attention}( \mathbf{Q}, \mathbf{K}, \mathbf{V}) = \text{softmax}\left(\frac{\mathbf{Q}\mathbf{K}^T}{\sqrt{d}}\right) \cdot \mathbf{V}, \text{where} \\
\mathbf{Q} = \mathbf{W}^{(i)}_Q \cdot  \varphi_i(z_t), \mathbf{K} = \mathbf{W}^{(i)}_K \cdot \phi(y), \mathbf{V} = \mathbf{W}^{(i)}_V \cdot \phi(y) .
\end{align}

$\varphi_i(z_t) \in \mathbb{R}^{N \times d^i_\epsilon}$ represents spatially flattened tokens of video latent, $\phi$ denotes the Clip text encoder, and $y$ is the input text prompt.
Motion speed control with fps is incorporated through an FPS embedder, which shares the same structure as the timestep embedder.
Specifically, the FPS or timestep is projected into an embedding vector using sinusoidal embedding.
This vector is then fed into a two-layer MLP to map the sinusoidal embedding to a learned embedding.
Subsequently, the timestep embedding and FPS embedding are fused via elementwise addition.
The fused embedding is finally added to the convolutional features to modulate the intermediate features.
 
\subsection{VideoCrafter1: Image-to-Video Model}
\label{subsec:t2v}
%--------------------------------------

\input{fig_tex/fig_img_branch}

\input{fig_tex/fig_enrich_emb}

Text prompts offer highly flexible control for content generation, but they primarily focus on semantic-level specifications rather than detailed appearance. In the I2V model, we aim to integrate an additional conditional input, \textit{i.e.}, image prompt, into the video diffusion model, which is expected to synthesize dynamic visual content based on the provided image.
For text-to-video diffusion models, the conditional text embedding space plays a crucial role in determining the visual content of the final output videos. To supply the video model with image information in a compatible manner, it is essential to project the image into a text-aligned embedding space. We propose learning such an embedding with rich details to enhance visual fidelity. Figure~\ref{fig:img_branch} illustrates the diagram of equipping the diffusion model with an image conditional branch.

\paragraph{Text-Aligned Rich Image Embedding.}
Since the text embedding is constructed using the pretrained CLIP~\cite{radford2021learning} text encoder, we employ its image encoder counterpart to extract the image features from the input image. Although the global semantic token $\mathbf{f}_{cls}$ from the CLIP image encoder is well-aligned with image captions, it primarily represents visual contents at a semantic level, while being less capable of capturing details. Inspired by existing visual conditioning works~\cite{shi2023instancebooth,ye2023ipadapter}, we utilize the full patch visual tokens $\mathbf{F}_{vis}=\{\mathbf{f}_i\}^K_{i=0}$ from the last layer of the CLIP image ViT~\cite{dosovitskiy2020image}, which are believed to encompass much richer information about the image.

To promote alignment with the text embedding, we utilize a learnable projection network $\mathcal{P}$ to transform $\mathbf{F}_{vis}$ into the target image embedding $\mathbf{F}_{img}=\mathcal{P}(\mathbf{F}_{vis})$, enabling the video model backbone to process the image feature efficiently. The text embedding $\mathbf{F}_{text}$ and image embedding $\mathbf{F}_{img}$ are then used to compute the U-Net intermediate features $\mathbf{F}_{in}$ via dual cross-attention layers:
\begin{equation}
\mathbf{F}_{out} = \text{Softmax}(\frac{\mathbf{Q} \mathbf{K}_{text}^{\top}}{\sqrt{d}})\mathbf{V}_{text} + \text{Softmax}(\frac{\mathbf{Q} \mathbf{K}_{img}^{\top}}{\sqrt{d}})\mathbf{V}_{img},
\end{equation}
where $\mathbf{Q}=\mathbf{F}_{in}\mathbf{W}_q$, $\mathbf{K}_{text}=\mathbf{F}_{text}\mathbf{W}_k$, $\mathbf{V}_{text}=\mathbf{F}_{text}\mathbf{W}_v$, and $\mathbf{K}_{img}=\mathbf{F}_{img}\mathbf{W}'_k$, $\mathbf{V}_{img}=\mathbf{F}_{img}\mathbf{W}'_v$ accordingly. 
Note that we use the same query for image cross-attention as for text cross-attention. Thus, only two parameter matrices $\mathbf{W}'_k$, $\mathbf{W}'_v$ are newly added for each cross-attention layer. 
Figure~\ref{fig:rich_emb} compares the visual fidelity of the generated videos conditioned on the global semantic token and our adopted rich visual tokens, respectively.

%% file: fig_tex/fig_frame_t2v.tex
\begin{figure*}[t]
  \centering
   \includegraphics[width=1.0\linewidth]{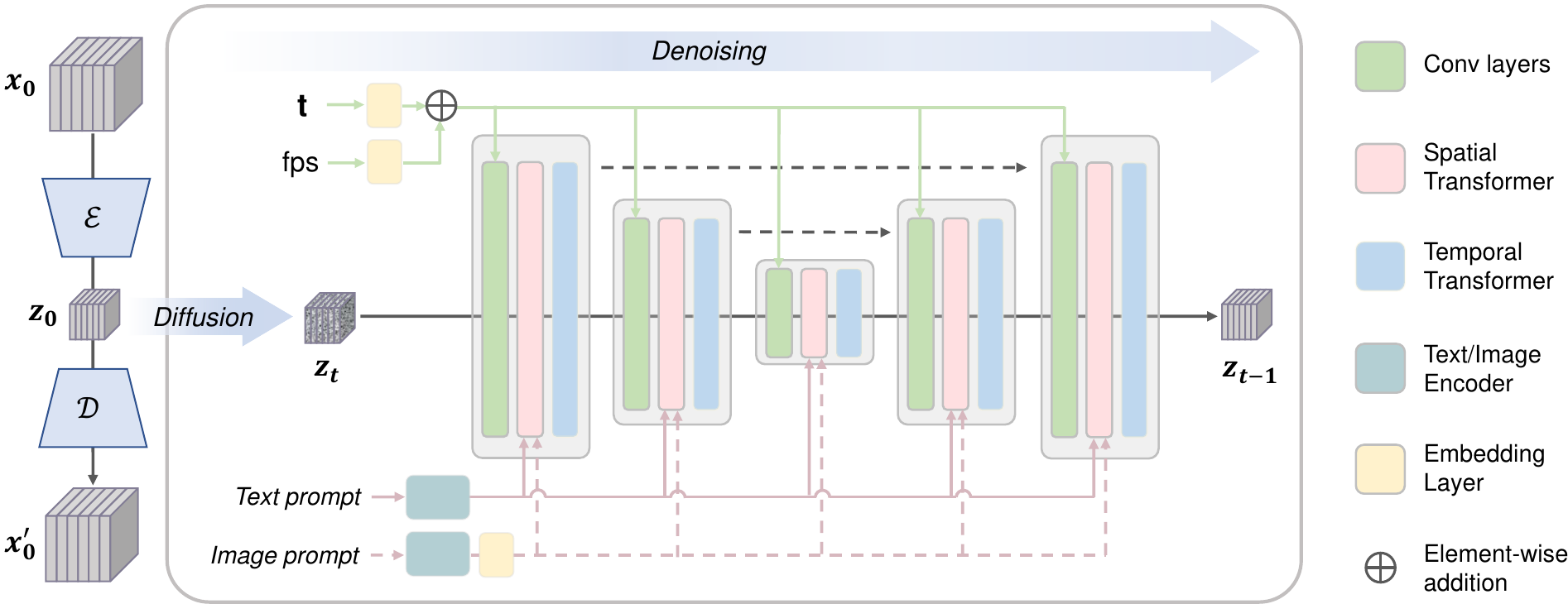}
   \caption{The framework of the video diffusion model in VideoCrafter1. We train the video UNet in the latent space of the auto-encoder. FPS is taken as a condition to control the motion speed of the generated video. For the T2V model, only the text prompt is fed into the spatial transformer via cross-attention, while for the I2V model, both the text and image prompts are taken as the inputs.}
   \label{fig:framework-t2v}
\end{figure*}

%% file: fig_tex/fig_img_branch.tex
\begin{figure}[!t]
    \centering
    \includegraphics[width=1\linewidth]{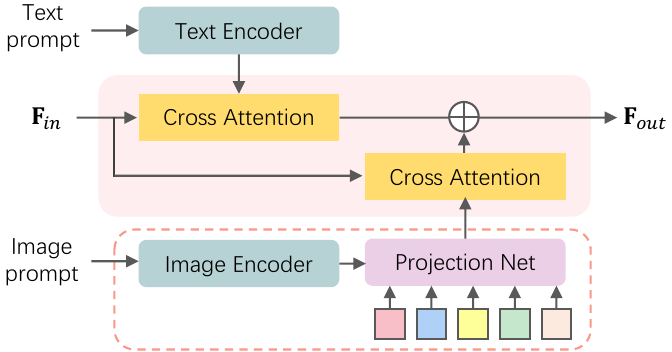}
    \caption{The diagram of image conditional branch. The U-Net backbone features $\mathbf{F}_{in}$ are processed with the text and image embeddings via a dual cross-attention layer, the output of which are fused as $\mathbf{F}_{out}$.}
    \label{fig:img_branch}
\end{figure}

%% file: fig_tex/fig_enrich_emb.tex
\begin{figure}[!t]
    \centering
    \includegraphics[width=1\linewidth]{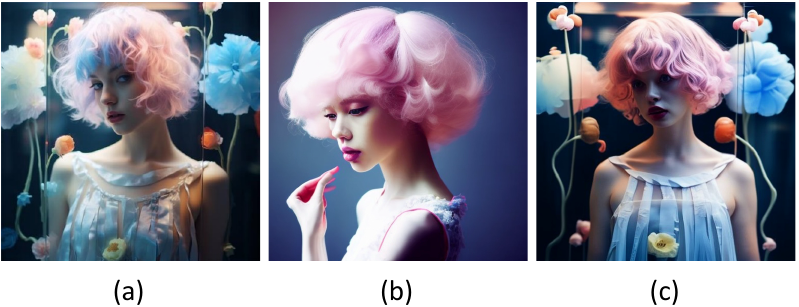}
    \caption{Image-conditioned text-to-video generation comparison. (a) Conditional image input. (b) Generation with the global semantic token conditioned. (c) Generation with the full patch visual tokens conditioned. The used text prompt is "a beautiful girl with colorful hair".}
    \label{fig:rich_emb}
\end{figure}

%% file: tex/4-experiment.tex
\input{fig_tex/eval}

\input{fig_tex/fig_user_study}

\input{fig_tex/fig_demo2}

\input{fig_tex/fig_videocrafter_version}

\input{fig_tex/fig_i2v_demo}

\section{Experiments}
\label{sec:exp}
%--------------------------------------

\subsection{Implementation Details}

\paragraph{Datasets.}
We employ an image and video joint training strategy for model training. The image dataset used is LAION COCO~\cite{LAION-COCO}, a large text-image dataset consisting of 600 million generated high-quality captions for publicly available web images. For video datasets, we utilize the publicly available WebVid-10M~\cite{Bain21}, a large-scale dataset of short videos with textual descriptions sourced from stock footage sites, offering diverse and rich content. Additionally, we compile a large-scale high-quality video dataset containing 10 million videos with resolutions greater than $1280 \times 720$  for the training of T2V models.

\paragraph{Training Scheme.}
To train the T2V model, we employ the training strategy used in Stable Diffusion, \textit{i.e.,} training from low resolution to high resolution.
We first train the video model extended from the image model at a resolution of $256\times 256$ for $80K$ iterations with a batch size of $256$.
Next, we resume from the $256\times 256$ model and finetune it with videos for $136K$ iterations at a resolution of $512\times 320$. The batch size is $128$.
Finally, we finetune the model for $45K$ iterations at a resolution of $1024 \times 576$. The batch size is $64$.
For the I2V model, we initially train the mapping from the image embedding to the embedding space used for the cross attention.
Subsequently, we fix the mappings of both text and image embeddings and finetune the video model for improved alignment.

\paragraph{Evaluation Metrics.}
We employ comprehensive metrics to assess video quality and the alignment between text and video using EvalCrafter~\cite{liu2023evalcrafter}, a benchmark for evaluating video generation models.
EvalCrafter conducts comparisons among our model, Gen-2, Pika Labs, and ModelScope, considering both quantitative metrics and user studies. We present the main results in Table~\ref{tab:overall} and Figure~\ref{fig:user_study}.
Our T2V model achieves the best visual quality and video quality among open-source models. Please refer to EvalCrafter for further details.
For qualitative evaluation, we provide several visual examples in Figures~\ref{fig:demo1}, ~\ref{fig:demo2}, and \ref{fig:i2v_comp} for illustration.

\paragraph{Relations to Floor33.}
We deploy the two open-source models on a Discord channel named Floor33, to allow users to explore the capability of the models online by just typing the prompt.  
We add an optional function, prompt extension, to enrich the information in the user's prompt. 
The discord channel can be accessed at \url{https://discord.gg/rrayYqZ4tf}.

\subsection{Performance Evaluation}
\label{subsec:evaluation}

\paragraph{Text-to-Video Results.}
We compare our T2V model with commercial models such as Gen-2 and Pika Labs, as well as the open-source model I2VGen-XL.
Since I2VGen-XL is an image-to-video model, we first generate an image using SDXL and then employ I2VGen-XL to create a video.
The results are displayed in Fig.~\ref{fig:demo1} and ~\ref{fig:demo2}.
As shown in Table~\ref{tab:overall}, our model outperforms open-source T2V models in terms of visual quality and text-alignment.
Our model encourages large object movements during training, resulting in more significant motion in the generated videos compared to other models. However, larger motions can sometimes introduce errors in temporal consistency. Fig.~\ref{fig:demo1} and ~\ref{fig:demo2} also demonstrate our superiority in visual quality and concept composition compared to open-source models.
The image quality of Zeroscope is subpar, as it sometimes fails to generate content or produces artifacts like repetitive grids.

Gen-2 and Pika Labs consistently generate videos with high aesthetic scores, and noise is suppressed in their results.
Nevertheless, Gen-2 occasionally struggles with concept composition, as seen in the two examples in Fig.~\ref{fig:demo2}, and its results are overly smooth.
Pika Labs exhibits the best text-alignment performance but does not always generate the correct style, such as in the second example in Fig.~\ref{fig:demo2}. 

We also compare the differences between VideoCrafter versions to verify our efforts. As shown in Fig.~\ref{fig:version} and Table.~\ref{tab:overall}, our method has a great process this year in both visual quality, text-video alignment, temporal consistency, and motion quality. We also find that our latest version~(23.10) has achieved the same quality as Pika Lab~\cite{pikalab}, which demonstrates the benefits of our training and datasets.

\paragraph{Image-to-Video Results.}
We evaluate our method against existing state-of-the-art image-to-video approaches, including two open-source conditional video diffusion models and two commercial product demos. VideoComposer~\cite{wang2023videocomposer} is a recently released model for compositional video generation that supports text-image-to-video synthesis. I2VGen-XL~\cite{I2VGen-XL} is an open-source image-to-video generation project. Pika~\cite{pikalab} and Gen-2~\cite{Gen2} are well-known text-to-video generation products developed by commercial companies, which also support image-to-video applications.

The visual comparison results are illustrated in Figure~\ref{fig:i2v_comp}. We observe that Pika, Gen-2, and our approach achieve relatively better visual fidelity to the conditional image than VideoComposer and I2V-XL. Although the first frame is almost identical to the input image, VideoComposer suffers from severe temporal inconsistency, where the subsequent frames transform into entirely different appearances. I2V-XL exhibits good temporal consistency and motion magnitude, but the appearance deviates significantly from the conditional image.
Pika achieves the best visual fidelity and temporal consistency; however, it generally presents very subtle motion magnitude. In contrast, Gen-2 can generate satisfying motion magnitude and visual fidelity, but its performance is not stable, \textit{i.e.}, it sometimes suffers from temporal drifting problems (as in the car case). 
Our I2V model demonstrates better performance in these cases, with good temporal consistency and motion magnitude, as well as acceptable visual fidelity.
However, our I2V model still has several limitations such as the successful rate, unsatisfactory facial artifacts, \textit{etc}, requiring further efforts for improvement.

%% file: fig_tex/eval.tex
\begin{table}[t]
\resizebox{\columnwidth}{!}{%
\begin{tabular}{lcccc}
\toprule
{} & Visual & Text-Video & Motion & Temporal \\ 
{} & Quality & Alignment & Quality & Consistency \\ \midrule

% \multicolumn{6}{c}{User Study}    \\ 
I2VGen-XL$^{\dag}$  & 55.23  & 47.22   & 59.41  & 59.31      \\ 
ZeroScope      & 56.37  & 46.18  & 54.26 & 61.19  \\ 
% VideoCrafter1  & 61.64   &  \textbf{66.76}  & 56.06  & 60.36 \\
% VideoCrafter1 & 59.53~(3) & {51.29~(3)}   & 51.97~(5)  & 56.36~(5)  \\ \hline
 PikaLab$^*$   & 63.52   & \textbf{54.11}   & 57.74  & 69.35    \\ 
 Gen2$^*$           & \textbf{67.35}   & 52.30  & \textbf{62.53}  & \textbf{69.71}    \\ 
\hline
% \multicolumn{6}{c}{User Study}    \\ 
VideoCrafter$^{23.04}$  & 46.88   & 41.56   & \textbf{56.24*}  & 55.78    \\ 
VideoCrafter$^{23.08}$ & 59.53   & 51.29   & 51.97  & 56.36   \\ 
VideoCrafter$^{23.10}$ & \textbf{61.64}   &  \textbf{66.76}   & 56.06     &  \textbf{60.36}   \\ 
 
\bottomrule
\end{tabular}
}
\vspace{-1em}
\caption{
Human-preference aligned results from four different aspects, with the rank of each aspect in the brackets. * indicated these models are not open-sourced. 
}
\vspace{-1em}
\label{tab:overall}
\end{table}

%% file: fig_tex/fig_user_study.tex
% Use figure* for multi-column figure
\begin{figure}[t]
    \centering
    \includegraphics[width=\linewidth]{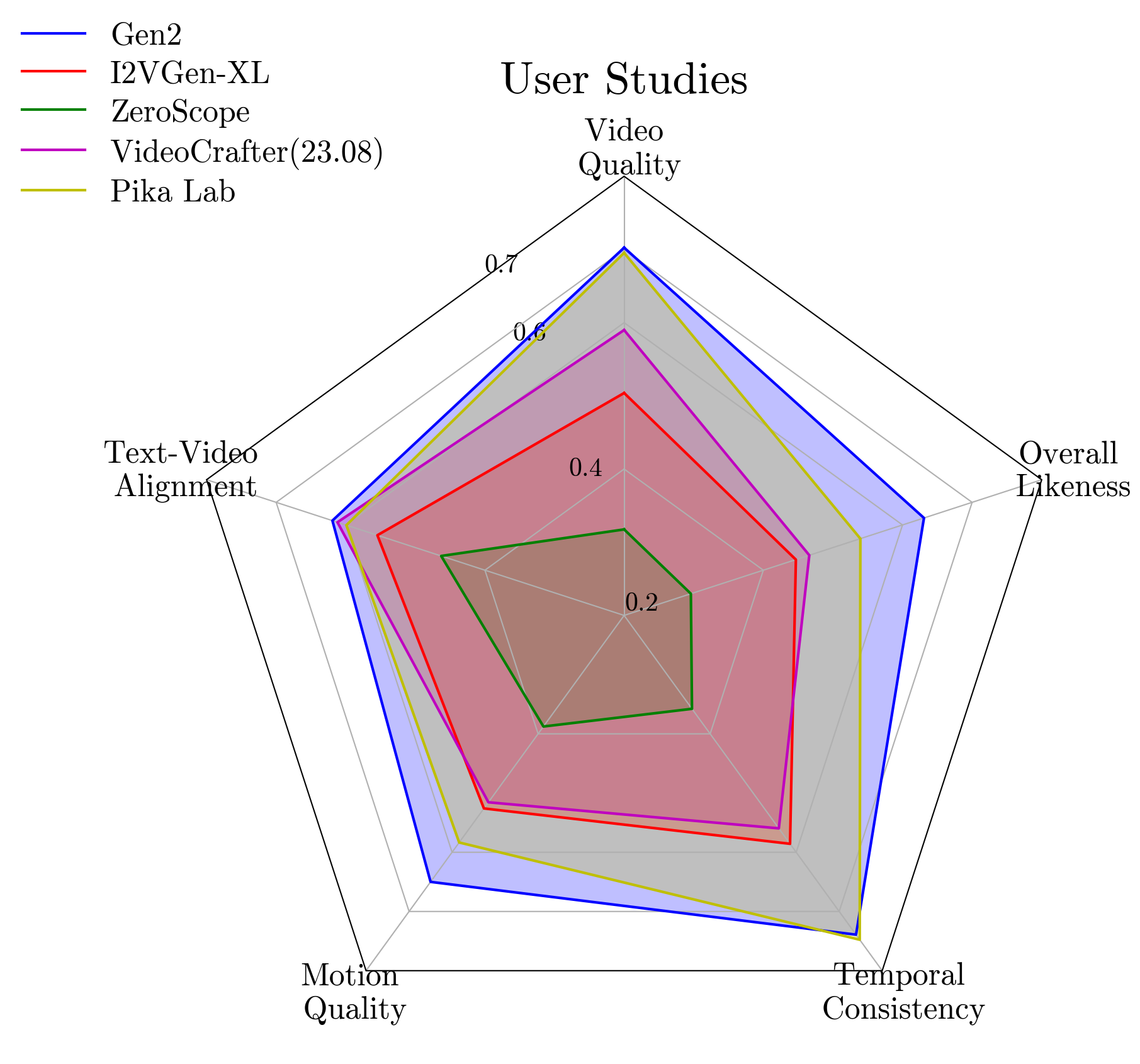}
    \vspace{-2em}
    \caption{The raw ratings from our user studies. }
    \vspace{-1em}
    \label{fig:user_study}
\end{figure}

%% file: fig_tex/fig_demo2.tex
\begin{figure*}[t]
  \centering
   \includegraphics[width=0.86\linewidth]{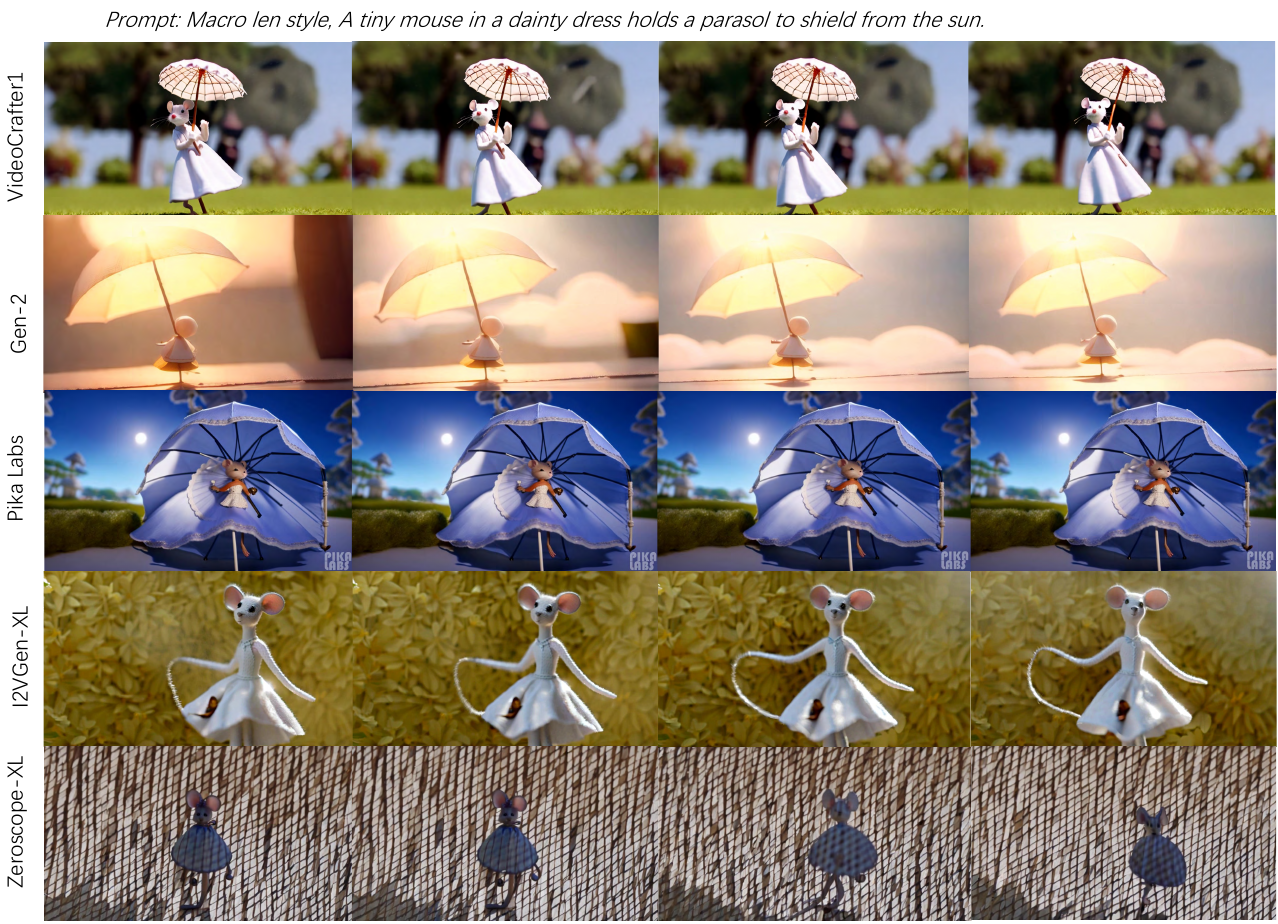}
   \includegraphics[width=0.86\linewidth]{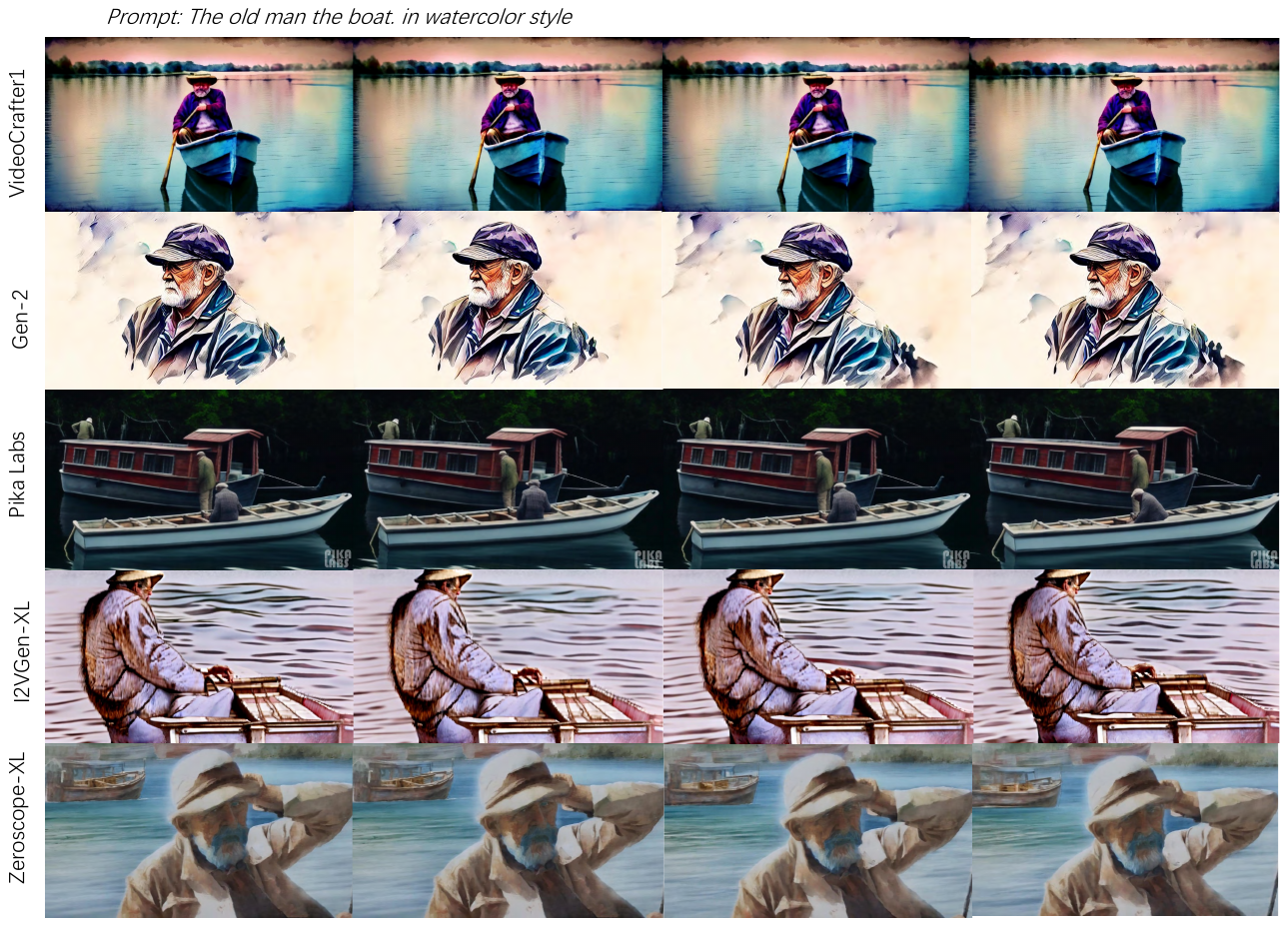}
   \caption{Visual comparison with Gen-2, Pika Labs, I2VGen-XL, and Zeroscope-XL.}
   \label{fig:demo2}
\end{figure*}

%% file: fig_tex/fig_videocrafter_version.tex
% Use figure* for multi-column figure
\begin{figure*}[tp]
    \centering
    \includegraphics[width=0.9\linewidth]{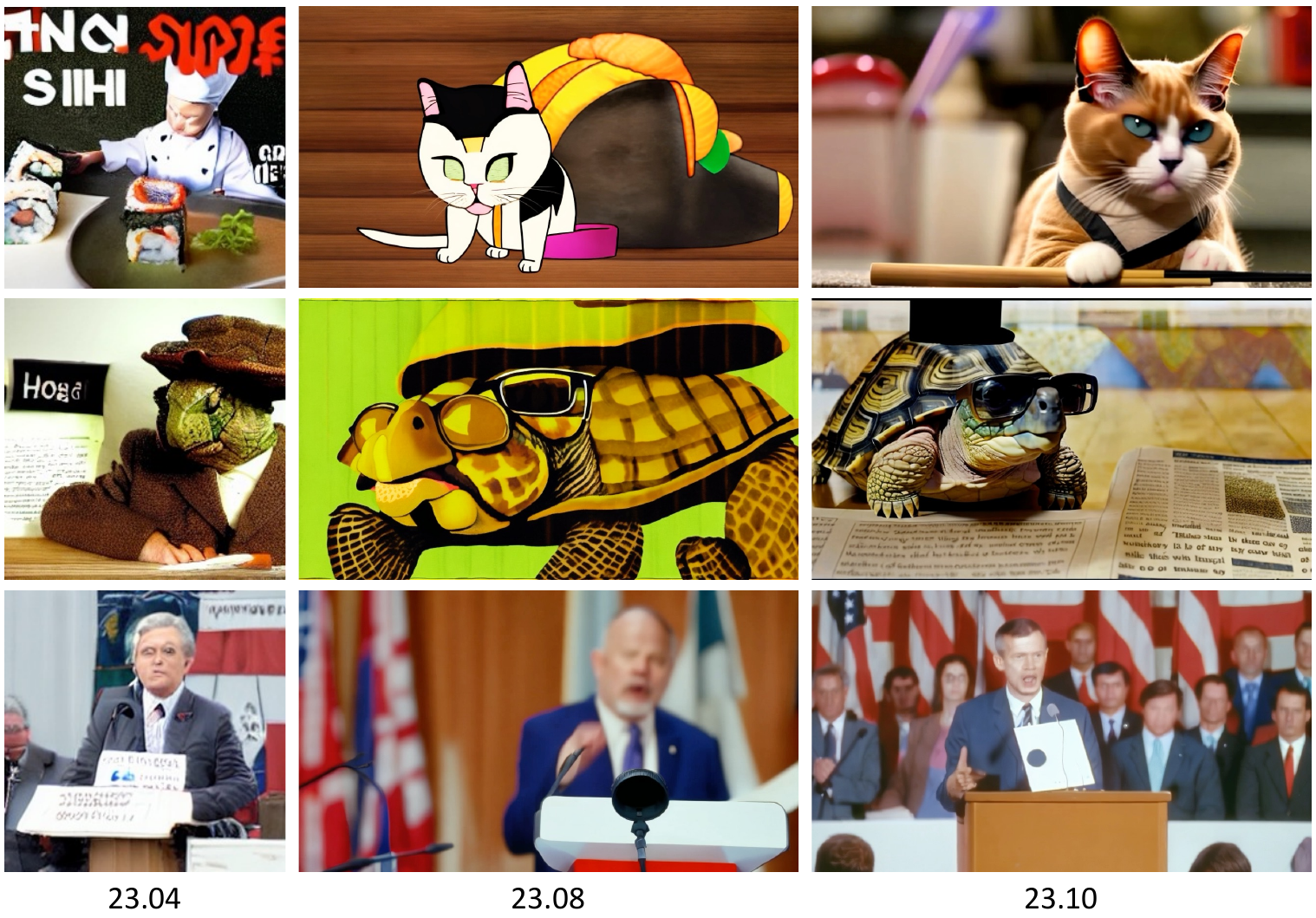}
    \vspace{-1em}
    \caption{The visual comparisons of the visual quality between different VideoCrafter text-to-video versions. The prompts are ``\textit{In Marvel movie style, supercute siamese cat as sushi chef}'', "\textit{A wise tortoise in a tweed hat and spectacles reads a newspaper, Howard Hodgkin style}" and ``\textit{hand-held camera, a politician giving a speech at a podium}'', respectively. The comparison video will be released on our Github.}
    \vspace{-1em}
    \label{fig:version}
\end{figure*}

%% file: fig_tex/fig_i2v_demo.tex
\begin{figure*}[!t]
    \centering
    \includegraphics[width=1\linewidth]{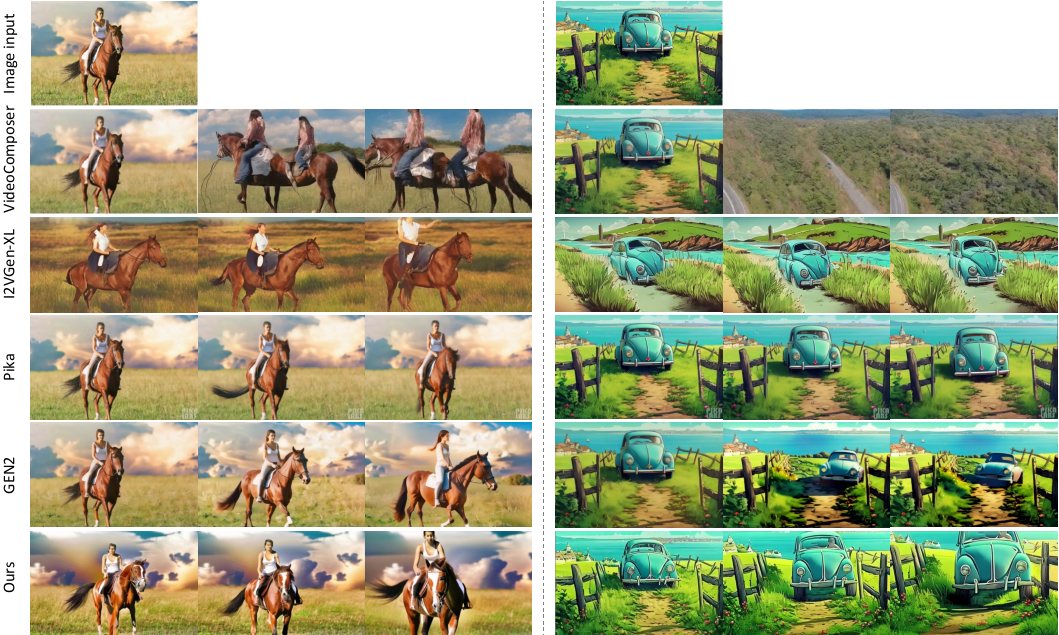}
    \caption{Visual comparisons with image-to-video approaches: VideoComposer, I2VGen-XL, Pika, Gen-2 and our I2V model.}
    \label{fig:i2v_comp}
\end{figure*}

%% file: tex/5-conclusion.tex
\section{Conclusion and Future Work}
\label{sub:conclusion}
%--------------------------------------

We introduce two diffusion models for video generation.
One is a text-to-video generation model capable of producing high-quality, high-resolution, cinematic-quality videos with a resolution of $1024 \times 576$. It offers the best quality among open-source T2V models.
The other is an image-to-video generation model, which is the first open-source generic I2V foundation model that can preserve the content and structure of the given reference image.

The existing open-source models merely represent the starting point.
Improvements in duration, resolution, and motion quality remain crucial for future developments.
Specifically, the current duration of the two models is limited to 2 seconds; extending this to a longer duration would be more beneficial.
This can be accomplished by training with additional frames and developing a frame-interpolation model. As for resolution, employing a spatial upscaling module or collaborating with ScaleCrafter~\cite{he2023scalecrafter} presents a promising strategy. Moreover, improvements in motion and visual quality can be achieved by utilizing higher-quality data.